\documentclass[10pt,twocolumn,letterpaper]{article}

\usepackage{cvpr}
\usepackage{graphicx}
\usepackage{amsmath}
\usepackage{amssymb}
\usepackage{booktabs}
\usepackage{verbatim}
\usepackage{enumitem}
\usepackage{makecell}
\usepackage{multirow}
\usepackage[linesnumbered,ruled,vlined]{algorithm2e}
\usepackage[pagebackref,colorlinks]{hyperref}

\usepackage[capitalize]{cleveref}
\crefname{section}{Sec.}{Secs.}
\Crefname{section}{Section}{Sections}
\Crefname{table}{Table}{Tables}
\crefname{table}{Tab.}{Tabs.}
\crefname{algorithm}{Alg.}{Algs.}

\newcommand{\tabref}[1]{Tab.~\ref{#1}}
\newcommand{\equref}[1]{Eq.~\ref{#1}}
\newcommand{\figref}[1]{Fig.~\ref{#1}}
\newcommand{\secref}[1]{Sec.~\ref{#1}}
\newcommand{\algref}[1]{Alg.~\ref{#1}}
\DeclareMathOperator*{\argmax}{argmax} 

\SetCommentSty{mycommfont}

\SetKwInput{KwInput}{Input}                
\SetKwInput{KwOutput}{Output}              

\begin{document}

\title{Dynamic Neural Network for Multi-Task Learning \\ Searching across Diverse Network Topologies}

\author{Wonhyeok Choi, Sunghoon Im\thanks{Corresponding author}\\
Department of Electrical Engineering \& Computer Science, DGIST, Daegu, Korea\\
{\tt\small \{smu06117, sunghoonim\}@dgist.ac.kr}
}
\maketitle

\begin{abstract}
\vspace{-2mm}
In this paper, we present a new MTL framework that searches for structures optimized for multiple tasks with diverse graph topologies and shares features among tasks.
We design a restricted DAG-based central network with read-in/read-out layers to build topologically diverse task-adaptive structures while limiting search space and time.
We search for a single optimized network that serves as multiple task adaptive sub-networks using our three-stage training process.
To make the network compact and discretized, we propose a flow-based reduction algorithm and a squeeze loss used in the training process.
We evaluate our optimized network on various public MTL datasets and show ours achieves state-of-the-art performance. 
An extensive ablation study experimentally validates the effectiveness of the sub-module and schemes in our framework.
\end{abstract}
\vspace{-5mm}

\section{Introduction}
\label{sec:intro}

Multi-task learning (MTL), which learns multiple tasks simultaneously with a single model has gained increasing attention~\cite{jou2016deep, bilen2016integrated, huang2015cross}.
MTL improves the generalization performance of tasks while limiting the total number of network parameters to a lower level by sharing representations across tasks.
However, as the number of tasks increases, it becomes more difficult for the model to learn the shared representations, and improper sharing between less related tasks causes negative transfers that sacrifice the performance of multiple tasks~\cite{kang2011learning, standley2020tasks}.
To mitigate the negative transfer in MTL, some works~\cite{chen2022task, misra2016cross, ruder2017sluice} separate the shared and task-specific parameters on the network.

\begin{figure*}
  \centering
  \begin{subfigure}{0.6\linewidth}
    \includegraphics[width=1.0\linewidth]{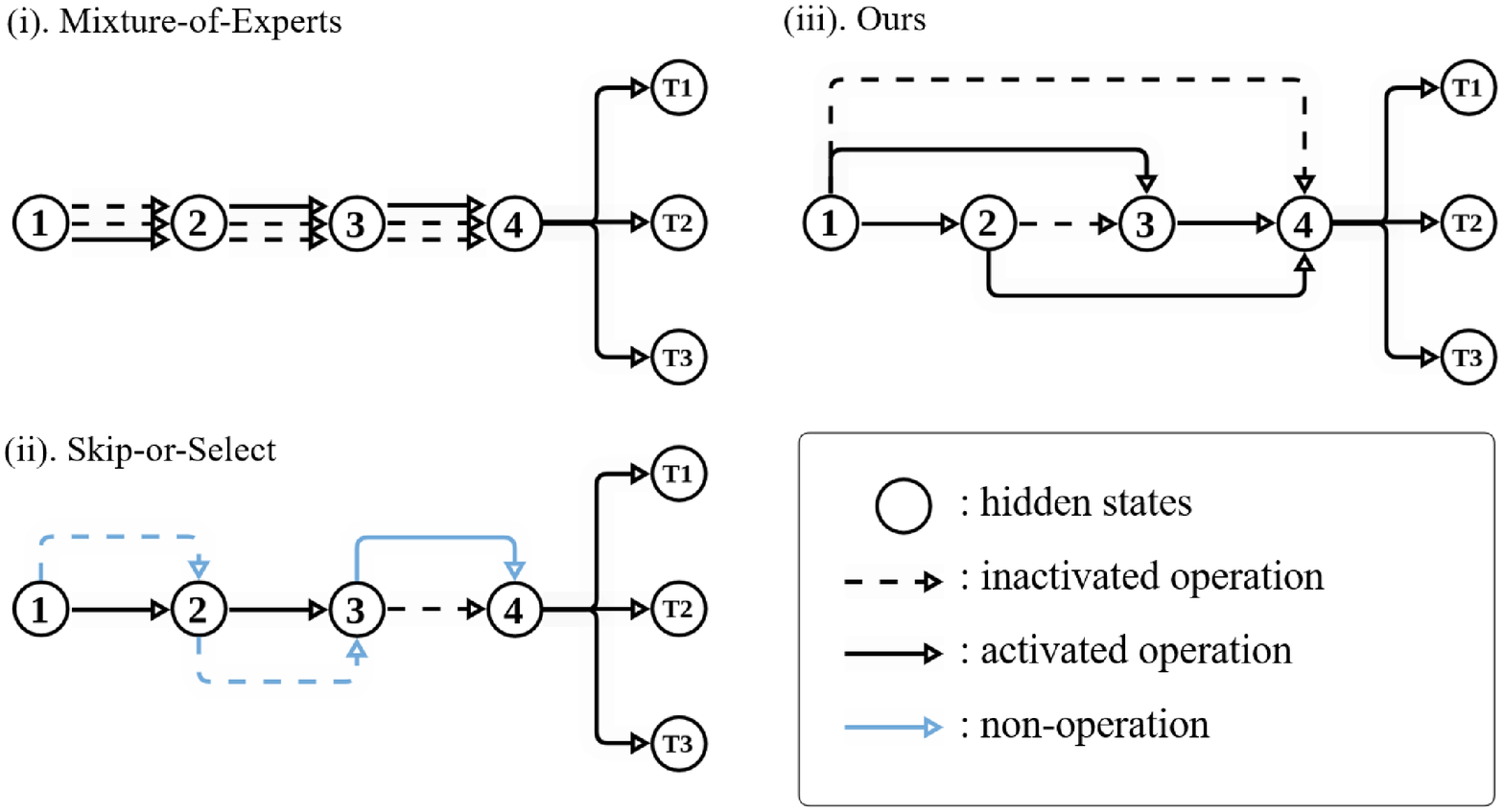}
    \caption{Graph representation of existing DNN-based methods (i).-(ii). for MTL and ours (iii).}
    \label{fig:short-a}
  \end{subfigure}
  \begin{subfigure}{0.32\linewidth}
    \includegraphics[width=1.0\linewidth]{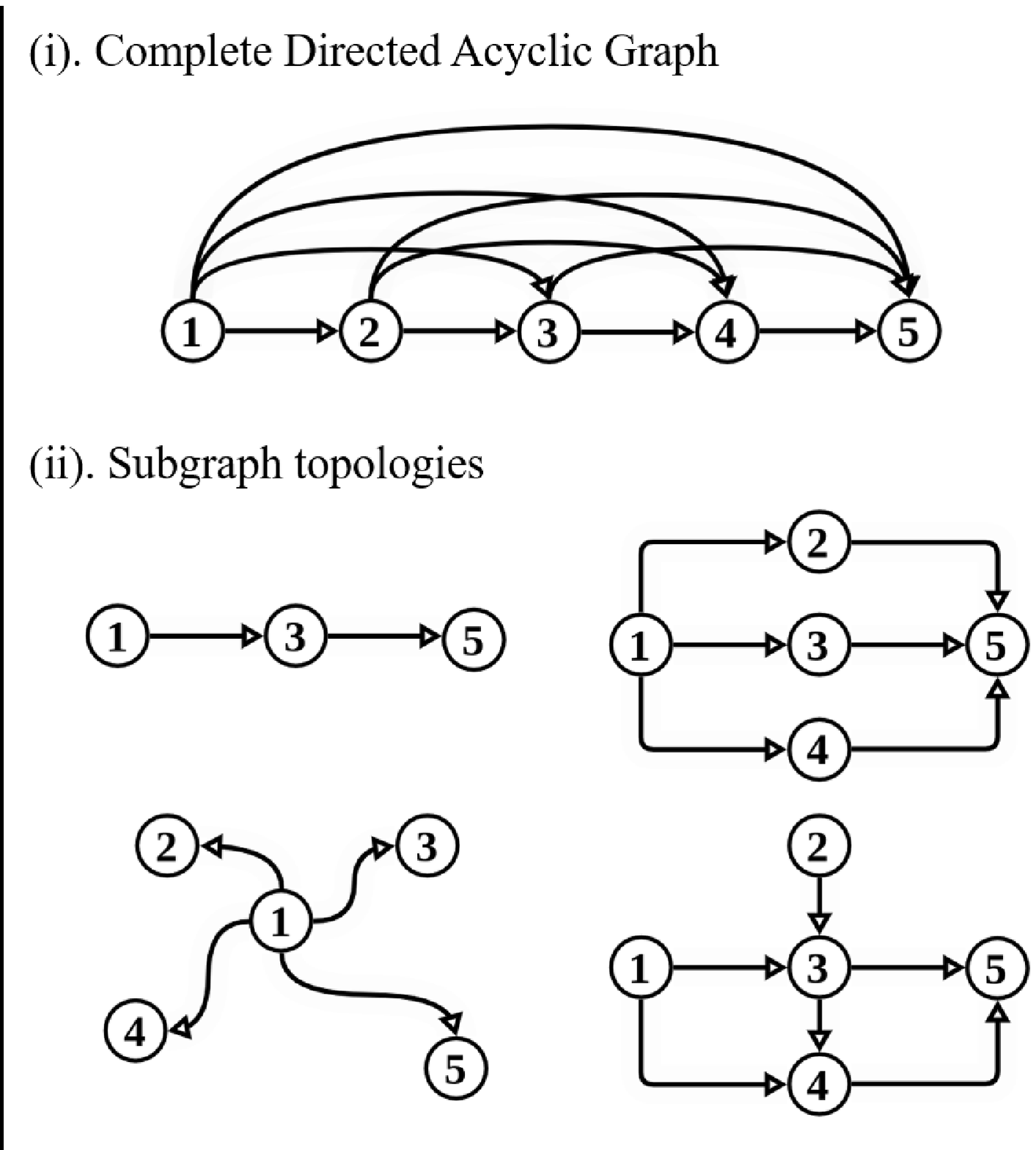}
    \caption{Topologies of DAG (i). and its sub-graph (ii).}
    \label{fig:short-b}
  \end{subfigure}
  \caption{\textbf{Graph representation of various neural networks.} (a) Graph representation of existing dynamic neural network for multitask learning and ours. (b) Topologies of a completed Directed Acyclic Graph (DAG) and the output sub-graph of DAG structure.}
  \label{fig:short}
\vspace{-10pt}
\end{figure*}

More recent works~\cite{sun2020adashare, raychaudhuri2022controllable, ma2019snr} have been proposed to dynamically control the ratio of shared parameters across tasks using a Dynamic Neural Network (DNN) to construct a task adaptive network.
These works mainly apply the cell-based architecture search~\cite{pham2018efficient, zoph2018learning, liu2018darts} for fast search times, so that the optimized sub-networks of each task consist of fixed or simple structures whose layers are simply branched, as shown in \figref{fig:short-a}.
They primarily focus on finding branching patterns in specific aspects of the architecture, and feature-sharing ratios across tasks.
However, exploring optimized structures in restricted network topologies has the potential to cause performance degradation in heterogeneous MTL scenarios due to unbalanced task complexity.

We present a new MTL framework searching for sub-network structures, optimized for each task across diverse network topologies in a single network.
To search the graph topologies from richer search space, we apply Directed Acyclic Graph (DAG) for the homo/heterogeneous MTL frameworks, inspired by the work in NAS~\cite{liu2018darts, pham2018efficient, zoph2016neural}.
The MTL in the DAG search space causes a scalability issue, where the number of parameters and search time increase quadratically as the number of hidden states increases.

To solve this problem, we design a restricted DAG-based central network with read-in/read-out layers that allow our MTL framework to search across diverse graph topologies while limiting the search space and search time.
Our flow-restriction eliminates the low-importance long skip connection among network structures for each task, and creates the required number of parameters from $O(N^2)$ to $O(N)$. 
The read-in layer is the layer that directly connects all the hidden states from the input state, and the read-out layer is the layer that connects all the hidden states to the last feature layer.
These are key to having various network topological representations, such as polytree structures, with early-exiting and multi-embedding.

Then, we optimize the central network to have compact task-adaptive sub-networks using a three-stage training procedure. 
To accomplish this, we propose a squeeze loss and a flow-based reduction algorithm.
The squeeze loss limits the upper bound on the number of parameters.
The reduction algorithm prunes the network based on the weighted adjacency matrix measured by the amount of information flow in each layer.
In the end, our MTL framework constructs a compact single network that serves as multiple task-specific networks with unique structures, such as chain, polytree, and parallel diverse topologies, as presented in \figref{fig:short-b}. 
It also dynamically controls the amount of sharing representation among tasks.

The experiments demonstrate that our framework successfully searches the task-adaptive network topologies of each task and leverages the knowledge among tasks to make a generalized feature.
The proposed method outperforms state-of-the-art methods on all common benchmark datasets for MTL.
Our contributions can be summarized as follows:
\begin{itemize}
\setlength\itemsep{0.05em}
    \item We present for the first time an MTL framework that searches both task-adaptive structures and sharing patterns among tasks. It achieves state-of-the-art performance on all public MTL datasets.
    \item We propose a new DAG-based central network composed of a flow restriction scheme and read-in/out layers, that has diverse graph topologies in a reasonably restricted search space.
    \item We introduce a new training procedure that optimizes the MTL framework for compactly constructing various task-specific sub-networks in a single network.
\end{itemize}

\section{Related Works}
\label{sec:rel}

\noindent\textbf{Neural Architecture Search (NAS)}
Neural Architecture Search is a method that automates neural architecture engineering~\cite{elsken2019neural}.
Early works~\cite{zoph2016neural, baker2016designing, zoph2018learning} use reinforcement learning based on rewarding the model accuracy of the generated architecture.
Alternative approaches~\cite{real2017large, suganuma2017genetic, miikkulainen2019evolving}  employ evolutionary algorithms to optimize both the neural architecture and its weights.
These methods search for an adequate neural architecture in a large discrete space.
Gradient-based NAS methods \cite{liu2018darts, xie2018snas, cai2018proxylessnas} of formulating operations in a differentiable search space are proposed to alleviate the scalability issues.
They generally use the convex combination from a set of operations instead of determining a single operation.
Most NAS approaches~\cite{liu2018darts, zoph2016neural, pham2018efficient} adopt the complete DAG as a search space, to find the architecture in the various network topologies.
However, DAG-based MTL frameworks have not been proposed, because of their considerably high computational demands.

\noindent\textbf{Multi-Task Learning (MTL)}
Multi-task learning in deep neural networks can be categorized into hard and soft parameter sharing types~\cite{ruder2017overview}.
Hard parameter sharing~\cite{jou2016deep, huang2015cross, bilen2016integrated}, also known as shared-bottom, is the most commonly used approach to MTL. 
This scheme improves generalization performance while reducing the computational cost of the network, by using shared hidden layers between all tasks.
However, it typically struggles with the negative transfer problem~\cite{kang2011learning, standley2020tasks} which degrades performance due to improper sharing between less relevant tasks.
\vspace{-2pt}

On the other hand, soft-parameter sharing~\cite{misra2016cross, ruder2017sluice} alleviate the negative transfer problem by changing the shared parameter ratio.
These approaches mitigate the negative transfer by flexibly modifying shared information, but they cannot maintain the computational advantage on the classic shared-bottom model.
Recently, advanced approaches have been proposed to adjust shared parameters using a dynamic neural network~\cite{sun2020adashare, maziarz2019flexible, raychaudhuri2022controllable, ma2019snr} and NAS~\cite{gao2020mtl}.

\noindent\textbf{NAS-style MTL}
MTL frameworks using a dynamic neural network (DNN) can be divided into two categories.
One employs the Mixture-of-Experts (MoE)~\cite{shazeer2017outrageously}, which is designed for conditional computation of per-sample, to MTL by determining the experts of each task~\cite{fernando2017pathnet, ma2019snr, maziarz2019flexible}. 
They have a fixed depth finalized task-specific sub-network, because they choose experts from a fixed number of modular layers. 
This causes a critical issue with task-balancing in the heterogeneous MTL.
The other adopts the skip-or-select policy to select task-specific blocks from the set of residual blocks~\cite{sun2020adashare} or a shared block per layer~\cite{raychaudhuri2022controllable, guo2020learning}. 
These methods only create a simple serial path in the finalized sub-network of each task, and a parallel link cannot be reproduced.
Moreover, they heuristically address the unbalanced task-wise complexity issues in the heterogenous MTL (\eg manually changing the balancing parameters based on the task complexity~\cite{sun2020adashare, raychaudhuri2022controllable}).
Thus, none of the aforementioned works focus on finding the optimal task-specific structure in the MTL scenario.


\begin{figure*}
  \centering
  \includegraphics[width=1.0\linewidth]{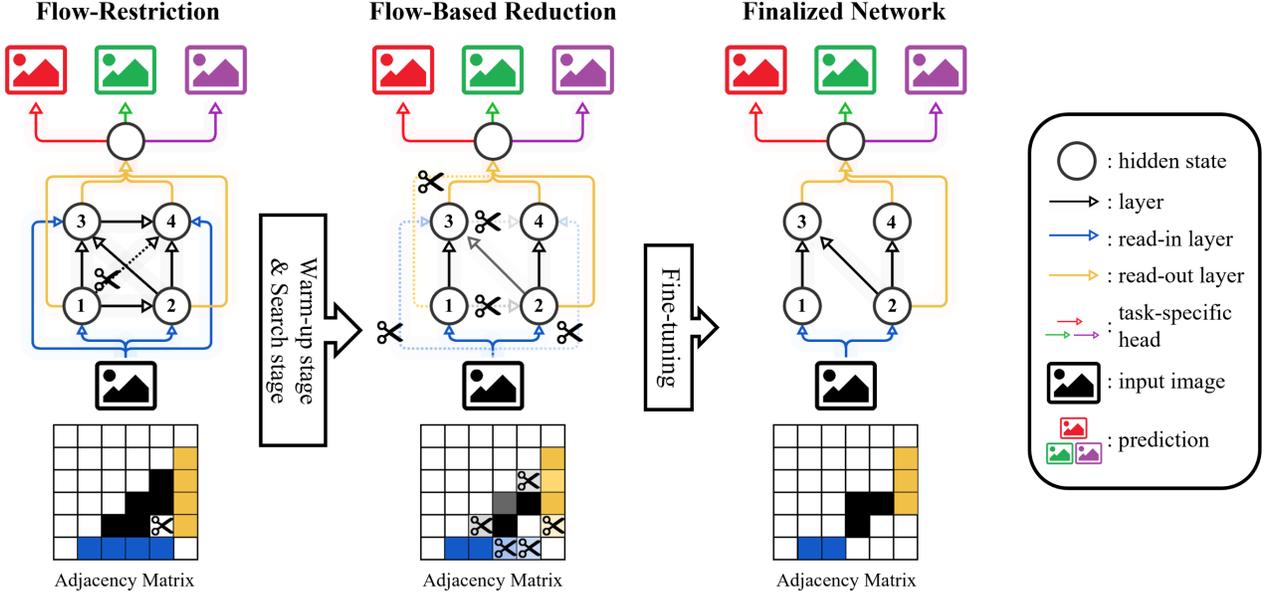}
  \caption{\textbf{Overall pipeline.} Our central network follows a DAG-based structure with read-in/out layers, and task-specific heads. The long skip connection is cut by our flow-restriction. Our framework with a 3-task MTL learning scenario consists of three stages including warm-up, search, and fine-tuning stages. The warm-up stage only learns the parameters of the main network $\Theta$ and task-specific weights. The search stage learns the upper-level parameters $\mathcal{A}, \mathcal{B}, \Gamma$, and task-specific weights. Then, flow-based reduction eliminates the low-importance edges from the network. The fine-tuning stage re-trains the network with the remaining important parameters.}
  \label{fig:main}
  \vspace{-1mm}
\end{figure*}

\section{Method}
\label{sec:method}
We describe our MTL framework, which searches for optimized network structures tailored to each task across diverse graph topologies, while limiting search time.
\secref{sec:search} describes the composition of the searchable space of our central network and our flow-restriction method for efficiently balancing the topological diversity of task-specific sub-networks and searching space.
\secref{sec:decision} introduces our mechanism to determine the task-adaptive sub-network in the central network and describes the overall training process and loss function.
The overall pipeline of our method is illustrated in \figref{fig:main}.

\subsection{The Central Network with Diverse Topologies}
\label{sec:search}

Our central network composes a graph $G = (V, E)$ with layers $E$ in which the $N$ hidden states $V = \{v_1,...,v_N\}$ are topologically sorted:
\begin{gather}
E = \{e_{ij}\}_{i,j \in \{1,...,N\}},~\text{where}~i < j,\\
e_{ij}(\text{ $\cdot$ } ;\theta_{ij}): \mathbb{R}^{N^{v_i}} \rightarrow \mathbb{R}^{N^{v_j}},
\label{eq:edge}
\end{gather}
where $e_{ij}$ is the operation that transfer the state $v_i$ to $v_j$ with the weight parameters $\theta_{ij} \in \Theta$, and $N^{v_k}$ is the number of the elements of hidden state $v_k$, respectively.
We adopt the DAG structure \cite{liu2018darts, pham2018efficient, zoph2016neural} for the network.
However, the optimized structure from DAG is searched from $2^{N(N-1)/2}$ network topologies, which are too large to be optimized in time.
To address the issue while maintaining diversity, we propose a flow-restriction and read-in/read-out layers.

\noindent\textbf{Flow-restriction} The flow-restriction eliminates the low-importance long skip connection among network structures for each task by restricting $j-i \leq M$ where $M$ is the flow constant.
Regulating the searchable edges in the graph reduces the required number of parameters and searching time from $O(N^2)$ to $O(N)$, but it sacrifices the diversity and capacity of the network topologies.

To explain the topological diversity and task capacity of sub-networks, we define the three components of network topology, as follows:
\begin{itemize}
    \item[1.] $\mathcal{D}(G) = \max(\{{\text{Distance}(v_i, v_j)}\}_{v_i,v_j \in V}),$
    \item[2.] $\mathcal{W}(G) = \max(\{\text{Out}_{v_i}\}_{v_i \in V}),$
    \item[3.] $\mathcal{S}(G_s, G) = |E_s|/|E|,$
\end{itemize}
where $\text{Out}_{v_i}$ is the out-degree of the vertex ${v_i}$ and $\text{Distance}(\cdot)$ is the operation that counts the number of layers (or edges) between two connected vertices.
The network depth $\mathcal{D}(G)$ is equal to the longest distance between two vertices in the graph $G$.
The network width $\mathcal{W}(G)$ is equal to the maximum value of the out-degrees of hidden states in the graph $G$.
The sparsity $\mathcal{S}(G_s, G)$ of the sub-graph $G_s$ is the ratio of finalized edges $|E_s|$ over entire edges $|E|$.
The first two components are measurements of the topological diversity of the finalized sub-network, while the last one is for the sub-network capacity.
While a complete DAG has the full range of depth and width components, the flow-restricted DAG has the properties of depth and width components as follows:
\begin{enumerate}[label=Property \arabic*.,itemindent=*]
  \item $\min(\{\mathcal{D}(G_s)\}_{G_s \subseteq G}) = \lceil (|V|/M) \rceil,$
  \item $\max(\{\mathcal{W}(G_s)\}_{G_s \subseteq G}) = M,$
\end{enumerate}
where $\{G_s\}$ is the entire sub-graph of $G$.
The min-depth property (Prop. 1) can cause the over-parameterized problem when the capacity of the task is extremely low. The max-width property (Prop. 2) directly sacrifices the diversity of network topologies in the search space.

\noindent\textbf{Read-in/Read-out layers} We design read-in/read-out layers to mitigate these problems.
The read-in layer embeds the input state $v_0$ into all hidden states $v_i\in V$ with task-specific weights $\alpha_{i}^{k} \in \mathcal{A}$ for all $K$ tasks $\mathcal{T}=\{T_k\}_{k \in \{1,...,K\}}$ as follows:

\begin{equation}
\label{eq:readin}
v_i^k = \sigma(\alpha_{i}^{k}) \cdot v_0,
\end{equation}
where $\sigma(\cdot)$ is the sigmoid function.
Then, the central network sequentially updates the hidden state $v_1^k$ to $v_{N}^k$ with the task-specific weights $\gamma_{ij}^k \in \Gamma$ that correspond to $e_{ij}^k$:
\begin{equation}
v_j^k = \frac{1}{\text{In}_{v_j^k}} \sum_{e_{ij} \in E} (\sigma(\gamma_{ij}^k) \cdot e_{ij}(v_i^k; \theta_{ij})),
\label{eq:mainnetwork}
\end{equation}
where $\text{In}_{v_j^k}$ is the in-degree of $v_j^k$. Note that $\Gamma$ is the adjacency matrix of graph $G$.
Finally, the read-out layer aggregates all hidden state features $\{v_i^k\}_{i \in \{{1,...,N}\}}$ with the task-specific weights $\beta_{i}^{k} \in \mathcal{B}$ and produces the last layer feature $v^k_L$ for each task $k$ as follows:
\begin{equation}
v^k_L = \sum_{i \in \{{1,...,N}\}} (\sigma(\beta_{i}^k) \cdot v_i^k).
\label{eq:readout}
\end{equation}
The final prediction $\hat{\mathbf{y}}_k$ for each task $T_k$ is computed by passing the aggregated features $v^k_L$ through the task-specific head $H^{k}(\cdot)$ as follows:
\begin{gather}
\begin{split}
\hat{\mathbf{y}}^{k} = H^k(v^k_L).
\end{split}
\label{eq:head}
\end{gather}

All upper-level parameters $\mathcal{A}, \mathcal{B}$, and $\Gamma$ are learnable parameters, and their learning process is described in~\secref{sec:decision}.
The read-in/read-out layers enable the optimized network to have a multi-input/output sub-network. The read-out layer aggregates all hidden states of the central network during the search stage, allowing a specific task to use the early hidden states to output predictions while ignoring the last few layers. These early-exit structures help alleviate the over-parameterized problem in simple tasks.

\subsection{Network Optimization and Training Procedure}
\label{sec:decision}
We describe the entire training process for our MTL framework, which consists of three stages, including warm-up, search, and fine-tuning stages.

\noindent\textbf{Warm-up stage} As with other gradient-based NAS, our framework has upper-level parameters that determine the network structure and parameters. 
This bilevel optimization with a complex objective function in an MTL setup makes the training process unstable. 
For better convergence, we initially train all network parameters across tasks for a few iterations.
We train the weight parameters of the central network $\Theta$ that shares all operations $E$ across tasks.
We fix all values of the upper-level parameters $\mathcal{A}, \mathcal{B}$, and $\Gamma$ as $0$, which becomes $0.5$ after the sigmoid function $\sigma(\cdot)$, and freeze them.
We train the network parameters $\Theta$ in \equref{eq:mainnetwork} with a task loss as follows:
\begin{equation}
\mathcal{L}_{task} = \sum_{k=0}^{K} \mathcal{L}_{T_k}(\hat{\mathbf{y}}_{T_k}, \mathbf{y}_{T_k}),
\end{equation}
where $\mathcal{L}_{T_k}$ is the task-specific loss, which is the unique loss function for each task.

\noindent\textbf{Search stage} After the warm-up stage, we unfreeze the upper-level parameters $\mathcal{A}, \mathcal{B}$, and $\Gamma$ and search the network topologies appropriate to each task.
We train all these parameters and network parameters $\Theta$ simultaneously by minimizing the task loss and the proposed squeeze loss $\mathcal{L}_{sq}$ as follows:
\begin{gather}
\mathcal{L}_{train} = \mathcal{L}_{task} + \lambda_{sq}\mathcal{L}_{sq},\\
\mathcal{L}_{sq} = \sum_{k=0}^{K}(\max(( \sum_{\gamma_{ij} \in \Gamma}(\sigma(\gamma_{ij})) - \kappa), 0)),
\end{gather}
where $\lambda_{sq}$ is the balancing hyperparameter, and $\kappa$ is a constant number called the budget, that directly reduces the sparsity of the central network.
This auxiliary loss is designed to encourage the model to save computational resources.

\setlength{\textfloatsep}{8pt}
\begin{algorithm}[t]
\DontPrintSemicolon
  \KwInput{$\Gamma \in \mathbb{R}^{N\times N}$, $\mathcal{A} \in \mathbb{R}^{N}$, $\mathcal{B} \in \mathbb{R}^{N}$}
  \KwOutput{$\hat\Gamma, \hat{\mathcal{A}}, \hat{\mathcal{B}}$ \tcp{Discretized params.}}
  \text{initialize zero matrix $\Psi, \hat\Psi \in \mathbb{R}^{(N+2) \times (N+2)}$}
  
  $N_{\alpha} = \argmax(\mathcal{A})$
  
  $N_{\beta} = \max(N_{\alpha} + 1, \argmax(\mathcal{B}))$
  
  $\Gamma[:N_{\alpha},:] \leftarrow 0$ \tcp{remove edges < $N_\alpha$}
  $\Gamma[:,N_{\beta}:] \leftarrow 0$ \tcp{remove edges > $N_\beta$}
  $\Psi[1:N, 1:N] \leftarrow \Gamma$ \tcp{merge $\Gamma, \mathcal{A}, \mathcal{B}$ into $\Psi$} 
  $\Psi[0, N_{\alpha}+1:N_{\beta}+1] \leftarrow \mathcal{A}[N_{\alpha}:N_{\beta}]$
  
  $\Psi[N_{\alpha}+1:N_{\beta}+1, N+1] \leftarrow \mathcal{B}[N_{\alpha}:N_{\beta}]$
  
  \While{True}
  {
    \text{initialize zero matrix $S \in \mathbb{R}^{(N+2) \times (N+2)}$}
    
    \For{$i \leftarrow 0~to~\{N+1\}$}
    {
        \For{$j \leftarrow 0~to~\{N+1\}$}
        {
          $S[i,j] \leftarrow \psi_{ij}
          \bigl(
          \frac{1}{\text{In}_{v_i}}  \sum_{\psi_{ki} \in \Psi}(\psi_{ki}) /
          {\sum_{\psi_{ik} \in \Psi}(\psi_{ik})}
          + \frac{1}{\text{Out}_{v_j}}  \sum_{\psi_{jk} \in \Psi}(\psi_{jk}) /
          \sum_{\psi_{kj} \in \Psi}(\psi_{kj})
          \bigl)$
        }
    }
    \text{$\psi_{ij} \leftarrow 0$, where $S[i,j]$ is nonzero min value}
    
    \If{graph builded from $\Psi$ is reachable}
    {
        $\hat\Psi \leftarrow \Psi$
    }
    \Else{
        $\hat\Psi[\hat\Psi > 0] \leftarrow 1$ \tcp{discretization}
        $\hat\Gamma \leftarrow \hat\Psi[1:N, 1:N]$ \tcp{split into $\Gamma, \mathcal{A}, \mathcal{B}$}
        
        $\hat{\mathcal{A}} \leftarrow \hat\Psi[0,1:N]$
        
        $\hat{\mathcal{B}} \leftarrow \hat\Psi[1:N, N+1]$
        
        return $\hat\Gamma, \hat{\mathcal{A}}, \hat{\mathcal{B}}$
    }
  }
\caption{Flow-based Reduction}
\label{alg:reduction}
\end{algorithm}

\noindent\textbf{Fine-tuning stage} Lastly, we perform a fine-tuning stage to construct a compact and discretized network structure using the trained upper-level parameters $\mathcal{A}, \mathcal{B}$, and $\Gamma$.
To do so, we design a flow-based reduction algorithm that allows the network to obtain high computational speed by omitting low-importance operations, as described in \algref{alg:reduction}.
It measures the amount of information flow of each layer $e_{ij}$ in the central network by calculating the ratio of edge weight with respect to other related edges weight.
Then, it sequentially removes the edge which has the lowest information flow.
\algref{alg:reduction} stops when the edge selected to be deleted is the only edge that can reach the graph.
We use the simple Depth-first search algorithm to check the reachability of $\hat\Gamma$ between hidden state $v_{N_\alpha}$ to $v_{N_\beta}$.
All the output $\hat{\mathcal{A}}, \hat{\mathcal{B}}, \hat\Gamma$ in \algref{alg:reduction}, which is the discretized binary adjacency matrix, represent the truncated task-adaptive sub-network. 
After the reduction, we fix the upper-level parameters and only re-train the network parameters $\Theta$, and we do not use the sigmoid function in \equref{eq:readin}-\ref{eq:readout}

\section{Experiments}
\label{sec:exp}


We first describe the experimental setup in \secref{sec:expsetup}. 
We compare our method to state-of-the-art MTL frameworks on various benchmark datasets for MTL in \secref{sec:results}.
We also conduct extensive experiments and ablation studies to validate our proposed method in \secref{sec:visualization}-\ref{sec:ablation}.

\subsection{Experimental Settings}
\label{sec:expsetup}

\noindent\textbf{Dataset} We use four public datasets for multi-task scenarios including
Omniglot~\cite{lake2015human}, NYU-v2~\cite{silberman2012indoor}, Cityscapes~\cite{Cordts2016Cityscapes}, and PASCAL-Context~\cite{mottaghi2014role}. We use these datasets, configured by previous MTL works~\cite{sun2020adashare,raychaudhuri2022controllable}, not their original sources.
\begin{itemize}
\setlength\itemsep{0.1em}
    \item \textbf{Omniglot} Omniglot is a classification dataset consisting of 50 different alphabets, and each of them consists of a number of characters with 20 handwritten images per character.
    \item \textbf{NYU-v2} NYU-v2 comprises images of indoor scenes, fully labeled for joint semantic segmentation, depth estimation, and surface normal estimation.
    \item \textbf{Cityscapes} Cityscapes dataset collected from urban driving scenes in European cities consists of two tasks: joint semantic segmentation and depth estimation.
    \item \textbf{PASCAL-Context} PASCAL-Context datasets contain PASCAL VOC 2010~\cite{silberman2012indoor} with semantic segmentation, human parts segmentation, and saliency maps, as well as additional annotations for surface normals and edge maps.
\end{itemize}

\noindent\textbf{Competitive methods} We compare the proposed framework with state-of-the-art methods \cite{meyerson2017beyond,ramachandran2018diversity,maziarz2019flexible,liang2018evolutionary,misra2016cross,ruder2017sluice,gao2019nddr,liu2019end,ahn2019deep,sun2020adashare,guo2020learning,raychaudhuri2022controllable} and various baselines including a single task and a shared-bottom.
The single-task baseline trains each task independently using a task-specific encoder and task-specific head for each task.
The shared-bottom baseline trains multiple tasks simultaneously with a shared encoder and separated task-specific heads.

We compare our method with MoE-based approaches, including Soft Ordering~\cite{meyerson2017beyond}, Routing~\cite{ramachandran2018diversity}, and Gumbel-Matrix~\cite{maziarz2019flexible}, as well as a NAS approach~\cite{liang2018evolutionary} on Omniglot datasets.
CMTR~\cite{liang2018evolutionary} can modify parameter count, similar to our method.
We compare our method with other soft-parameter sharing methods including Cross-Stitch~\cite{misra2016cross}, Sluice network~\cite{ruder2017sluice}, and NDDR-CNN~\cite{gao2019nddr} and the dynamic neural network (DNN)-based methods including MTAN~\cite{liu2019end}, DEN~\cite{ahn2019deep}, and Adashare~\cite{sun2020adashare} for the other three datasets.
We provide the evaluation results of two recent works, LTB~\cite{guo2020learning} and PHN~\cite{raychaudhuri2022controllable} for PASCAL-Context datasets because only the results are reported in their papers, but no source codes are provided.

\noindent\textbf{Multi-task scenarios} We set up multi-task scenarios with the combination of several tasks out of a total of seven tasks, including classification $\mathcal{T}_{cls}$, semantic segmentation $\mathcal{T}_{sem}$, depth estimation $\mathcal{T}_{dep}$, surface normal prediction $\mathcal{T}_{norm}$, human-part segmentation $\mathcal{T}_{part}$, saliency detection $\mathcal{T}_{sal}$, and edge detection $\mathcal{T}_{edge}$.
We follow the MTL setup in~\cite{sun2020adashare} for three datasets including Omniglot, NYU-v2, and cityscapes, and~\cite{raychaudhuri2022controllable} for PASCAL-Context.
We simulate a homogeneous MTL scenario of a 20-way classification task in a multi-task setup using Omniglot datasets by following \cite{meyerson2017beyond}.
Each task predicts a class of characters in a single alphabet set.
We use the other three datasets for heterogeneous MTL.
We set three tasks including segmentation, depth estimation, and normal estimation for NYU-v2 and two with segmentation, depth estimation for Cityscapes.
We set five tasks $\mathcal{T}_{sem}$, $\mathcal{T}_{part}$, $\mathcal{T}_{norm}$, $\mathcal{T}_{sal}$, and $\mathcal{T}_{edge}$ as used in~\cite{raychaudhuri2022controllable} for PASCAL-Context datasets.

\begin{table}[t]
\centering
\begin{tabular}{c|c|c}

\toprule
Method  & Test Acc. (\%) & $\#$ of Param $\downarrow$ \\ 
\midrule
Soft Ordering~\cite{meyerson2017beyond}   & 66.59  & \textbf{0.27}  \\
CMTR~\cite{liang2018evolutionary}         & 87.19  & -  \\
MoE~\cite{ramachandran2018diversity}      & 92.19  & 9.08  \\
Gumbel-Matrix~\cite{maziarz2019flexible}  & 93.52  & 9.08  \\
\midrule
Single Task & 93.48  & 20.00 \\
Shared Bottom  & 93.25  & 1.00 \\
\midrule
Ours ($M = 3$)    & 94.99           & 0.91\\ 
Ours ($M = 5$)    & \textbf{95.71}  & 1.37\\
Ours ($M = 7$)    & 95.68           & 1.46\\ 
\bottomrule

\end{tabular}
\caption{Evaluation on \textbf{Omniglot datasets}.}
\label{tab:omniglot}
\end{table}

\begin{table}[t]
\centering
\resizebox{\linewidth}{!}{
\begin{tabular}{c|ccc|c|c}
\toprule
Method & $\Delta_{\mathcal{T}_{sem}} \uparrow$ & $\Delta_{\mathcal{T}_{norm}} \uparrow$ & $\Delta_{\mathcal{T}_{dep}} \uparrow$ & $\Delta_{\mathcal{T}} \uparrow$ & $\#$ of Param $\downarrow$ \\
\midrule
Single-Task                              &  0.0            &  0.0            &  0.0              &  0.0       & 3.00         \\
Shared Bottom                            & -7.6            & +7.5            & +5.2              & +1.7       & \textbf{1.00}\\
Cross-Stitch~\cite{misra2016cross}       & -4.9            & +4.2            & +4.7              & +1.3       & 3.00         \\
Sluice~\cite{ruder2017sluice}            & -8.4            & +2.9            & +4.1              & -0.5       & 3.00         \\
NDDR-CNN~\cite{gao2019nddr}              & -15.0           & +2.9            & -3.5              & -5.2       & 3.15         \\
MTAN~\cite{liu2019end}                   & -4.2            & +8.7            & +3.8              & +2.7       & 3.11         \\
DEN~\cite{ahn2019deep}                   & -9.9            & +1.7            & -35.2             & -14.5      & 1.12         \\
AdaShare~\cite{sun2020adashare}          & +8.8            & +7.9            & +10.1             & +8.9       & \textbf{1.00}\\
\midrule
Ours ($M = 5$) &         +11.9    &         +7.9    &          +8.8     &  +9.5            & 1.04 \\
Ours ($M = 7$) & \textbf{+13.4}   & \textbf{+9.2}   &         +10.7     & \textbf{+11.1}   & 1.31 \\
Ours ($M = 9$) &         +13.2    &         +9.0    & \textbf{+10.9}    & +11.0            & 1.63 \\
\bottomrule
\end{tabular}
}
\caption{Evaluation on \textbf{NYU-v2 datasets}.}
\label{tab:nyu}
\end{table}

\noindent\textbf{Evaluation metrics} We follow the common evaluation metrics utilized in the competitive methods.
We use an accuracy metric for the classification task.
The semantic segmentation task is measured by mean Intersection over Union (mIoU) and pixel accuracy.
We use the mean absolute and mean relative errors, and relative difference as the percentage of $ \delta = \max ( {\hat{\mathbf{d}}} / {\mathbf{d}}, {\mathbf{d} / \hat{\mathbf{d}}} )$ within thresholds $1.25^{\{1,2,3\}}$ for the depth estimation task.
For the evaluation of the PASCAL-Context datasets, we follow the same metrics used in~\cite{raychaudhuri2022controllable} for all tasks.
As reported in \cite{sun2020adashare}, we report a single relative performance $\Delta_{\mathcal{T}_i}$ in \tabref{tab:omniglot}-\ref{tab:pascal} for each task $\mathcal{T}_i$ with respect to the single-task baseline, which defined as:
\begin{equation}
\label{eq:rel_performance1}
\Delta_{\mathcal{T}_i} = {\frac{100}{|\mathcal{M}|}} \sum_{j=0}^{|\mathcal{M}|}(-1)^{l_j}{\frac{(\mathcal{M}_{\mathcal{T}_i,j} - \mathcal{M}_{\mathcal{T}_i,j}^{single})}{\mathcal{M}_{\mathcal{T}_i,j}^{single}}}, 
\end{equation}
where $\mathcal{M}_{\mathcal{T}_i,j}$ and $\mathcal{M}_{\mathcal{T}_i,j}^{single}$ are the $j$-th metric of $i$-th task $\mathcal{T}_i$ from each method and the single task baseline, respectively.
The constant $l_j$ is 1 if a lower value represents better for the metric $\mathcal{M}_{\mathcal{T}_i,j}$ and 0 otherwise.
The averaged relative performance for all tasks $\mathcal{T}$ is defined as:

\begin{gather}
\label{eq:rel_performance2}
\Delta_{\mathcal{T}} = \frac{1}{|\mathcal{T}|} \sum_{i=1}^{|\mathcal{T}|} \Delta_{\mathcal{T}_i}.
\end{gather}
The absolute task performance for all metrics is reported in the supplementary material.

\noindent\textbf{Network and training details} 
For our central network, we set 8 hidden states, the same as the existing MoE-based works~\cite{ramachandran2018diversity, maziarz2019flexible} and use the same classification head for Omniglot datasets.
We set 12 hidden states, the same as the VGG-16~\cite{simonyan2014very}, except for the max-pooled state, and use the Deeplab-v2~\cite{chen2017deeplab} decoder structure as all task heads for all the other datasets, respectively.
We use the Adam~\cite{kingma2014adam} optimizer to update both upper-level parameters and network parameters.
We use cross-entropy loss for semantic segmentation and L2 loss for the other tasks.
For a fair comparison, we train our central network from scratch without pre-training for all experiments.
We describe more details on the network structure and hyperparameter settings in the supplementary material.

\begin{table}[t]
\centering
\resizebox{\linewidth}{!}{%
\begin{tabular}{c|cc|c|c}
\toprule
Method & $\Delta_{\mathcal{T}_{sem}} \uparrow$ & $\Delta_{\mathcal{T}_{dep}} \uparrow$ & $\Delta_{\mathcal{T}} \uparrow$ & $\#$ of Param $\downarrow$ \\
\midrule
Single-Task                              &  0.0            &  0.0            &  0.0              & 2.00          \\
Shared Bottom                            & -3.7            & -0.5            & -2.1              & 1.00          \\
Cross-Stitch~\cite{misra2016cross}       & -0.1            & \textbf{+5.8}   & +2.8              & 2.00          \\
Sluice~\cite{ruder2017sluice}            & -0.8            & +4.0            & +1.6              & 2.00          \\
NDDR-CNN~\cite{gao2019nddr}              & +1.3            & +3.3            & +2.3              & 2.07          \\
MTAN~\cite{liu2019end}                   & +0.5            & +4.8            & +2.7              & 2.41          \\
DEN~\cite{ahn2019deep}                   & -3.1            & -1.6            & -2.4              & 1.12          \\
AdaShare~\cite{sun2020adashare}          & +1.8            & +3.8            & +2.8              & 1.00          \\
\midrule
Ours  ($M = 5$)                          & +3.5            & +3.9            & +3.7              & \textbf{0.96} \\
Ours  ($M = 7$)                          & +7.5            & +3.1            & +5.3              & 1.16          \\
Ours  ($M = 9$)                          & \textbf{+8.3}   & +4.8            & \textbf{+6.6}     & 1.31          \\

\bottomrule
\end{tabular}%
}
\caption{Evaluation on \textbf{Cityscapes datasets}.}
\label{tab:cityscapes}
\end{table}

\begin{table}[t]
\centering
\resizebox{\linewidth}{!}{%
\begin{tabular}{c|ccccc|c|c}
\toprule
Method & $\Delta_{\mathcal{T}_{sem}} \uparrow$ & $\Delta_{\mathcal{T}_{part}} \uparrow$  & $\Delta_{\mathcal{T}_{sal}} \uparrow$  & $\Delta_{\mathcal{T}_{norm}} \downarrow$  & $\Delta_{\mathcal{T}_{edge}} \uparrow$ & $\Delta_{\mathcal{T}} \uparrow$ & $\#$ of Param $\downarrow$ \\
\midrule
Single-Task                             &  0.0  &  0.0          &  0.0 &  \textbf{0.0}  &  0.0  &  0.0     & 5.00          \\
Shared Bottom                           & -6.6  & -0.7          & -3.4 & -14.3          &  0.0  & -5.0     & \textbf{1.00} \\
Cross-Stitch~\cite{misra2016cross}      & -1.3  & +3.6          & -0.2 & -1.4           &  0.0  & +0.1     & 5.00          \\
Sluice~\cite{ruder2017sluice}           & -1.6  & -1.2          & -0.5 & -2.9           & -6.0  & -2.4     & 5.00          \\
NDDR-CNN~\cite{gao2019nddr}             & -1.1  & -2.6          &  0.0 & -5.0           &  0.0  & -1.7     & 5.61          \\
MTAN~\cite{liu2019end}                  & -3.6  & -0.7          & -0.3 & -5.0           & -6.0  & -3.1     & 5.21          \\
AdaShare~\cite{sun2020adashare}         & -1.4  & \textbf{+4.0} & -0.5 & -0.7           &  0.0  & +0.3     & \textbf{1.00} \\
LTB~\cite{guo2020learning}              & -6.9  & -1.9          & +0.2 & -1.4           &  0.0  & -2.0     & 3.19          \\
PHN~\cite{raychaudhuri2022controllable} & -6.6  & -1.6          & -1.0 &  \textbf{0.0}  &  0.0  & -1.8     & 2.51          \\
\midrule
Ours $(M=5)$     & -0.3          & +3.4  &         +0.9  &           0.0  & \textbf{0.0}  &         +0.8  & 1.93               \\
Ours $(M=7)$     & \textbf{0.0}  & -0.2  &         +1.7  &  \textbf{+1.4} & \textbf{0.0}  &         +0.6  & 1.91               \\
Ours $(M=9)$     & \textbf{0.0}  & +3.6  & \textbf{+1.8} &  \textbf{+1.4} & \textbf{0.0}  & \textbf{+1.4} & 2.31               \\
\bottomrule
\end{tabular}
}
\caption{Evaluation on \textbf{PASCAL-Context datasets}.}
\label{tab:pascal}
\end{table}


\subsection{Comparison to State-of-the-art Methods}
\label{sec:results}
We report the performance of the proposed method with different flow constants $M$ and compare it with state-of-the-art methods in \tabref{tab:omniglot}-\ref{tab:pascal} with four different MTL scenarios.
\tabref{tab:omniglot} shows that our framework with any flow constant $M$ outperforms all the competitive methods for the homogeneous MTL scenario with Omniglot datasets.
Ours has a similar number of parameters to the shared-bottom baseline.
All the other experiments for heterogeneous MTL scenarios in \tabref{tab:nyu}-\ref{tab:pascal} show that our frameworks achieve the best performance among all state-of-the-art works.
Even with the flow constant $M=5$, our model outperforms AdaShare~\cite{sun2020adashare} for both the NYU-v2 and Cityscapes datasets, while keeping almost the same number of parameters (NYU-v2: 1.00 vs. 1.04 and Cityscapes 1.00 vs. 0.96).
With the flow constant $M=7,9$, our method outperforms all the baselines by a large margin.
The results from the PASCAL-Context datasets in \tabref{tab:pascal} show that all baselines suffer from negative transfer in several tasks, as the number of tasks increases.
Only Adashare and Cross Stitch slightly outperform the single-task baseline (see the performance $\Delta_{\mathcal{T}}$).
On the other hand, ours with $M=9$ achieves the best performance without any negative transfers for all tasks.


\begin{figure}
  \centering
  \includegraphics[width=1.0\linewidth]{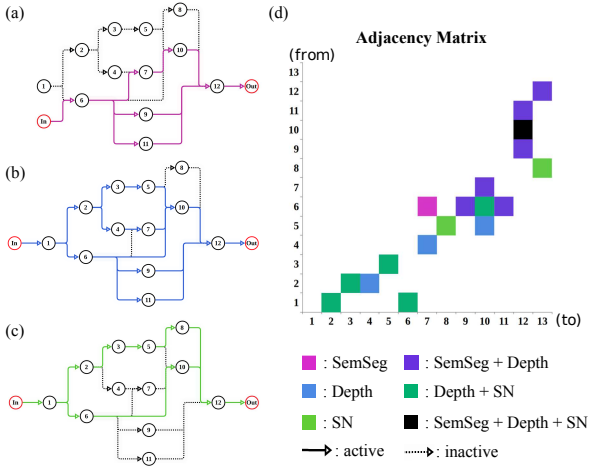}
  \caption{\textbf{Graph Representation of Task-adaptive Sub-network} The finalized sub-network topologies ($M=7$) trained with NYU-v2 datasets is illustrated as graph. (a-c) The task-adaptive sub-network of semantic segmentation, depth estimation, and surface normal, respectively. (d) The adjacency matrix where color represents the discretized value for the activated edge of each task.}
  \label{fig:graphvis}
\vspace{-1pt}
\end{figure}

\begin{table}[t]
\centering
\begin{tabular}{c|ccc}
\toprule
Task & $\mathcal{D}$ & $\mathcal{W}$ & $\mathcal{S}$\\
\midrule
Semantic Seg. & 5 & 3 & 0.103\\
Depth  & 7 & 3 & 0.192\\
Surface normal     & 7 & 2 & 0.128\\
\bottomrule
\end{tabular}
\caption{Topologies analysis on \textbf{NYU-v2 datasets}.}
\label{tab:topology_analysis}
\end{table}


Interestingly, the required parameters of the search space increase almost in proportion to the increase of the flow constant, but there is no significant difference in the number of parameters of the finalized networks.
For example, the required number of parameters for the network with the flow constant $M=3,5,7$ is 2.77, 4.23, and 5.38, respectively.
This demonstrates that the proposed flow-based reduction algorithm is effective in removing low-relative parameters while maintaining performance.
Specifically, we observe that the total performance of our framework with $M=7$ is slightly better than the $M=9$ setup in \tabref{tab:nyu} despite its smaller architecture search space.
To investigate this, we further analyze the tendency in performance and computational complexity with respect to the flow constant in \secref{sec:performance}.

\subsection{Analysis of Topologies and Task Correlation}
\label{sec:visualization}
To demonstrate the effectiveness of the proposed learning mechanism, we visualize our finalized sub-network topologies in \figref{fig:graphvis}-(a-c) and the adjacency matrix for NYU-v2 3-task learning in \figref{fig:graphvis}-(d).
We also analyze the diversity and capacity of task-adaptive network topologies in \tabref{tab:topology_analysis} with network depth $\mathcal{D}$, width $\mathcal{W}$, and sparsity $\mathcal{S}$ described in~\secref{sec:search}.
These analyses provide three key observations on our task-adaptive network and the correlation among tasks.

First, \textit{the tasks of segmentation and surface normal hardly share network parameters}. 
Various task-sharing patterns are configured at the edge, but there is only one sharing layer between the two tasks.
This experiment shows a low relationship between tasks, as it is widely known that the proportion of shared parameters between tasks indicates task correlation~\cite{sun2020adashare, raychaudhuri2022controllable, maziarz2019flexible}.

Second, \textit{long skip connections are mostly lost}.
The length of the longest skip connection in the finalized network is 5, and the number of these connections is 2 out of 18 layers, even with the flow constant of 7.
This phenomenon is observed not only in NYU-v2 datasets but also in the other MTL datasets.
This can be evidence that the proposed flow-restriction reduces search time while maintaining high performance even by eliminating the long skip connection in the DAG-based central network.

Lastly, \textit{Depth estimation task requires more network resources than segmentation and surface normal estimation tasks}.
We analyze the network topologies of the finalized sub-network of each task in NYU-v2 datasets using three components defined in \secref{sec:method}.
The depth $\mathcal{D}$ and width $\mathcal{W}$ of the sub-network increase in the order of semantic segmentation, surface normal prediction, and depth estimation tasks.
Likewise, the sparsity $\mathcal{S}$ of the depth network is the highest. 
This experiment shows that the depth network is the task that requires the most network resources.

\subsection{Performance w.r.t. Flow-resctriction}
\label{sec:performance}
\begin{figure}[t]
  \centering
  \includegraphics[width=1.0\linewidth]{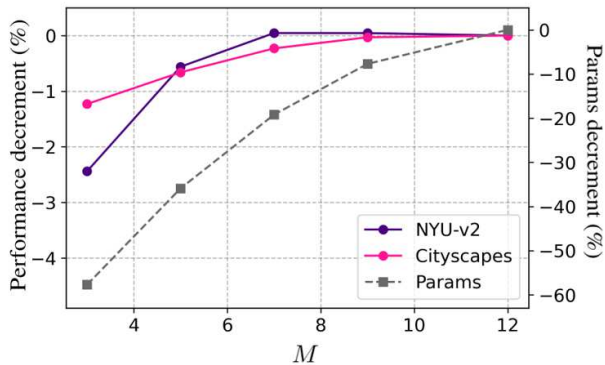}
  \caption{\textbf{Model performance with respect to the proposed flow-restriction.} We plot the degradation ratio of the performance (left y-axis) and parameter (right y-axis) by changing the flow constant $M$.
  We measure the final averaged task performance with NYU-v2 and Cityscapes datasets marked by purple and pink circle markers, respectively. We also measure the number of parameters marked by gray square markers.}
  \label{fig:chart1}
\vspace{-2pt}
\end{figure}

We analyze performance and computational complexity with respect to the flow constant $M$ for the NYU-v2 and Cityscapes datasets.
We report the rate of performance degradation with respect to the complete DAG search space in \figref{fig:chart1}.
We observe that the reduction rate of the final performance does not exceed 3\% even with considerably lower flow constants.
The performance is saturated at a flow constant of about 7 or more.
This means that the proposed method optimizes the task adaptive sub-network regardless of the size of the network search space, if it satisfies the minimum required size.

\begin{figure}[t]
  \centering
  \includegraphics[width=0.95\linewidth]{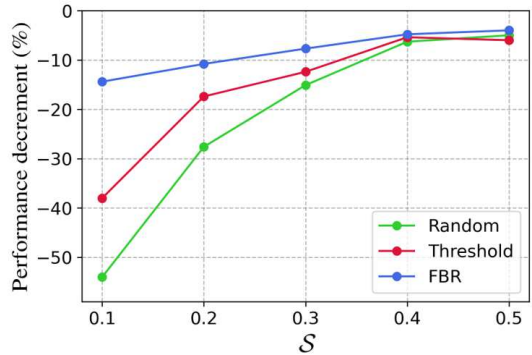}
  \caption{\textbf{Model Performance with respect to the network sparsity.} We plot the performance degradation rate by changing network sparsity. We compare our flow-based reduction algorithm to two other schemes; random selection and thresholding.}
  \label{fig:chart2}
\end{figure}

To demonstrate the effectiveness of our flow-based reduction (FBR) algorithm, we compare it to two other reduction algorithms (random and threshold) for Cityscapes datasets in \figref{fig:chart2}.
The random reduction literally removes the edges randomly, and the thresholding method sequentially removes the edge which has the lowest value in the adjacency matrix $\Gamma$.
We measure the rate of performance degradation of the pruned network of each reduction algorithm with respect to the non-reduced network while changing the sparsity $\mathcal{S}$.
Note that our method automatically determines the sparsity, so for this experiment only, we add a termination condition that stops the network search when a certain sparsity is met.
The results show that the proposed flow-reduction method retains performance even with a low sparsity rate.
This means that our method efficiently prunes the low-related edge of the network compared to the other methods.


\subsection{Ablation Study on Proposed Modules}
\label{sec:ablation}
We conduct ablation studies on the four key components of our framework; the flow-restriction, read-in/out layers, flow-based reduction, and squeeze loss.
We report the relative task performance and the number of parameters of the finalized network with/without the components in \tabref{tab:ablation}.
The results show that our framework, including all components, achieves the lowest number of parameters and the second-best performance.
Our method without flow-based reduction achieves the best performance.
However, the finalized network from this setup has about a five-times larger number of parameters than ours because the network has never been pruned in a training process.
This demonstrates that our restricted DAG-based central network is optimized to build compact task-adaptive sub-networks with performance close to the optimized sub-network from a complete DAG-based network.

\begin{table}[t]
\vspace{0pt}
\centering
\resizebox{\linewidth}{!}{%
\begin{tabular}{c|ccc|c|c}
\toprule
Method      & $\Delta_{\mathcal{T}_{sem}} \uparrow $ & $\Delta_{\mathcal{T}_{dep}} \uparrow $ & $\Delta_{\mathcal{T}_{norm}} \uparrow $ & $\Delta_{\mathcal{T}}$ & $\#$ of Param $\downarrow$ \\
\midrule
Ours (M=7)              & +13.4            & \textbf{+9.2}     & +10.7             & +11.1            & \textbf{1.31}\\
\midrule
w/o flow-restriction     & +13.2            & \textbf{+9.2}     & +10.4             & +11.0            & 1.80\\
w/o read-in/out          & +11.7            & +8.3              & +10.4             & +10.1            & 1.43\\
w/o flow-based reduction & \textbf{+14.2}   & \textbf{+9.2}     & \textbf{+11.1}    & \textbf{+11.5}   & 6.50\\
w/o $\mathcal{L}_{sq}$   & +13.2            & +8.8              & +10.7             & +10.9            & 1.38\\          
\bottomrule
\end{tabular}
}
\caption{\textbf{Ablation study on the proposed modules (NYU-v2).}}
\label{tab:ablation}
\vspace{-1pt}
\end{table}

\vspace{-1pt}
\section{Conclusions}
\label{sec:conclusion}
In this paper, we present a new MTL framework to search for task-adaptive network structures across diverse network topologies in a single network.
We propose flow restriction to solve the scalability issue in a complete DAG search space while maintaining the diverse network topological representation of the DAG search space by adopting read-in/out layers.
We also introduce a flow-based reduction algorithm that prunes the network efficiently while maintaining overall task performance and squeeze loss, limiting the upper bound on the number of network parameters.
The extensive experiments demonstrate that the sub-module and schemes of our framework efficiently improve both the performance and compactness of the network.
Our method compactly constructs various task-specific sub-networks in a single network and achieves the best performance among all the competitive methods on four MTL benchmark datasets.

\section*{Acknowledgement}
This work was supported by the National Research Foundation of Korea (NRF) grant funded by the Korea government (MSIT) (No. RS-2023-00210908).

\clearpage
{\small
\bibliographystyle{ieee_fullname}

}

\clearpage
\appendix
\section{Implementation Details}
\noindent\textbf{Central Network Architecture}
We set the first 12 hidden states, the same as the VGG-16~\cite{simonyan2014very}, except for the max-pooled states as:

\begin{table}[h]
\centering
\begin{tabular}{c|c}
\toprule
\textbf{State} & \textbf{Shape}\\
\midrule
$v_0$ (image state) & B, 3, H, W\\
\midrule
$v_1$ & B, 64, H, W\\
$v_2$ & B, 64, H, W\\
$v_3$ & B, 128, H//2, W//2\\
$v_4$ & B, 128, H//2, W//2\\
$v_5$ & B, 256, H//4, W//4\\
$v_6$ & B, 256, H//4, W//4\\
$v_7$ & B, 256, H//4, W//4\\
$v_8$ & B, 512, H//8, W//8\\
$v_9$ & B, 512, H//8, W//8\\
$v_{10}$ & B, 512, H//8, W//8\\
$v_{11}$ & B, 512, H//16, W//16\\
$v_{12}$ & B, 512, H//16, W//16\\
\midrule
$v_{13}$ (read-out state) & B, 512, H//16, W//16\\
\bottomrule
\end{tabular}
\caption{\textbf{Shape of all hidden states}}
\label{tab:hidden}
\vspace{-3pt}
\end{table}

\noindent where shapes of states are represented as (batch size, number of channels, height, and width).
Then, we link the states with edges as a block that consists of sequential operations as follows:
\begin{table}[h]
\centering
\begin{tabular}{|c|}
\toprule
$e_{ij}: v_i \rightarrow v_j$\\
\midrule
\text{conv3x3($C_{v_i}, C_{v_j}$, padding = 1, stride = 1)}, \\
\text{BatchNorm($C_{v_j}$)}, \\
\text{ReLU(),}\\
\text{Maxpool(kernel size = $H_{v_j} // H_{v_i}$)}\\
\bottomrule
\end{tabular}
\caption{\textbf{The operation block of $e_{ij}$}}
\label{block}
\vspace{-8pt}
\end{table}


\begin{figure}
  \centering
  \includegraphics[width=0.95\linewidth]{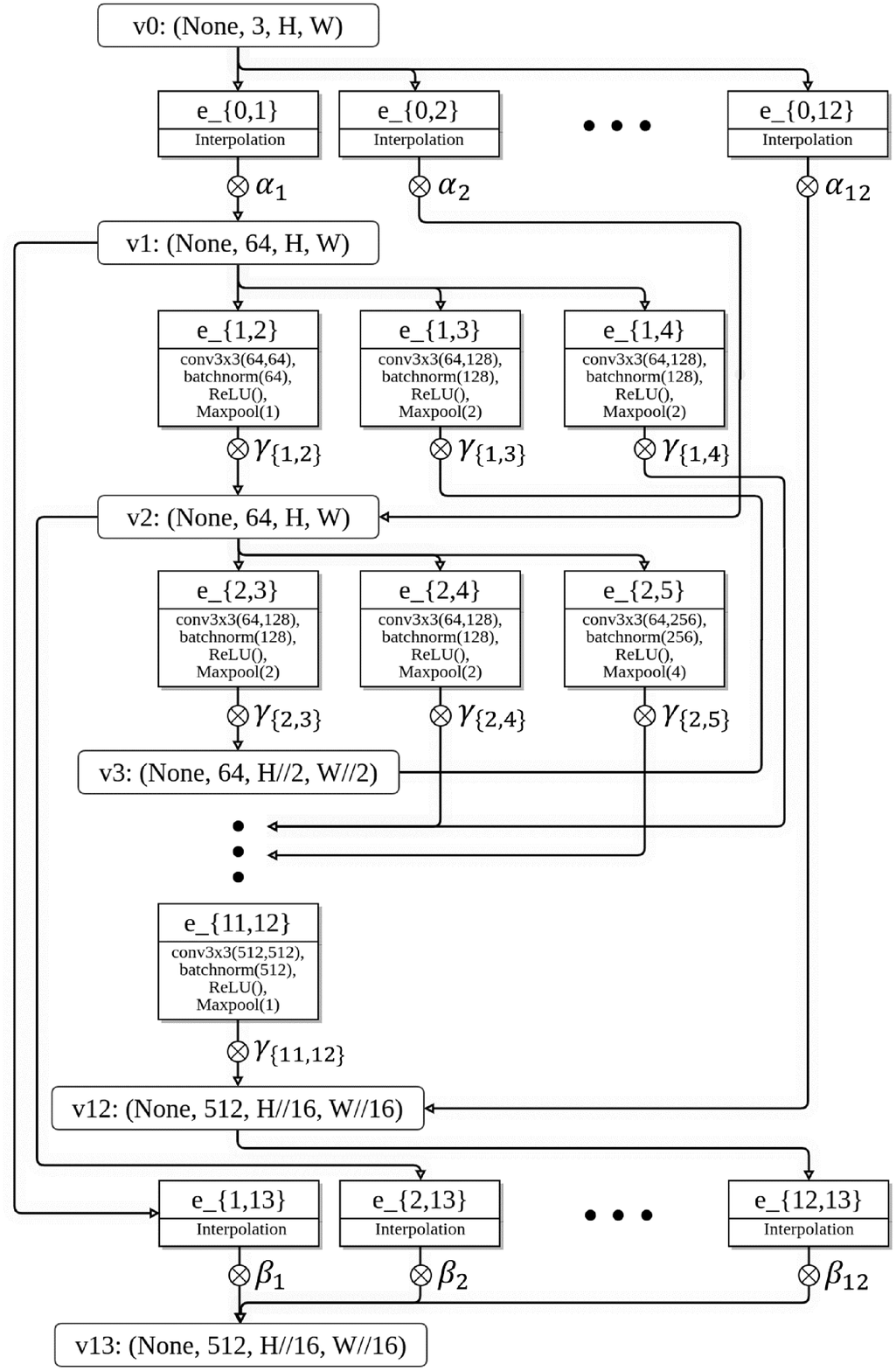}
  \caption{\textbf{Central network configuration}} 
  \label{fig:central-config}
  \vspace{-10pt}
\end{figure}


\noindent where $C_{v_i}$ is the number of channels of $v_i$, and $H_{v_i}$ is the height of $v_i$.
We illustrate the overall structure of the central network with $M=3$ in \figref{fig:central-config}. The read-in layer embeds the interpolated feature into all hidden states $v_1,v_2,...,v_{12}$ with $\alpha_i \in \mathcal{A}$. Then, the network sequentially updates the hidden states with task-specific weight $\gamma_{ij} \in \Gamma$ that corresponds to $e_{ij}$. Lastly, the read-out layer extracts the weighted sum of interpolated hidden states with $\beta_i \in \mathcal{B}$.

\begin{figure}
  \centering
  \includegraphics[width=0.95\linewidth]{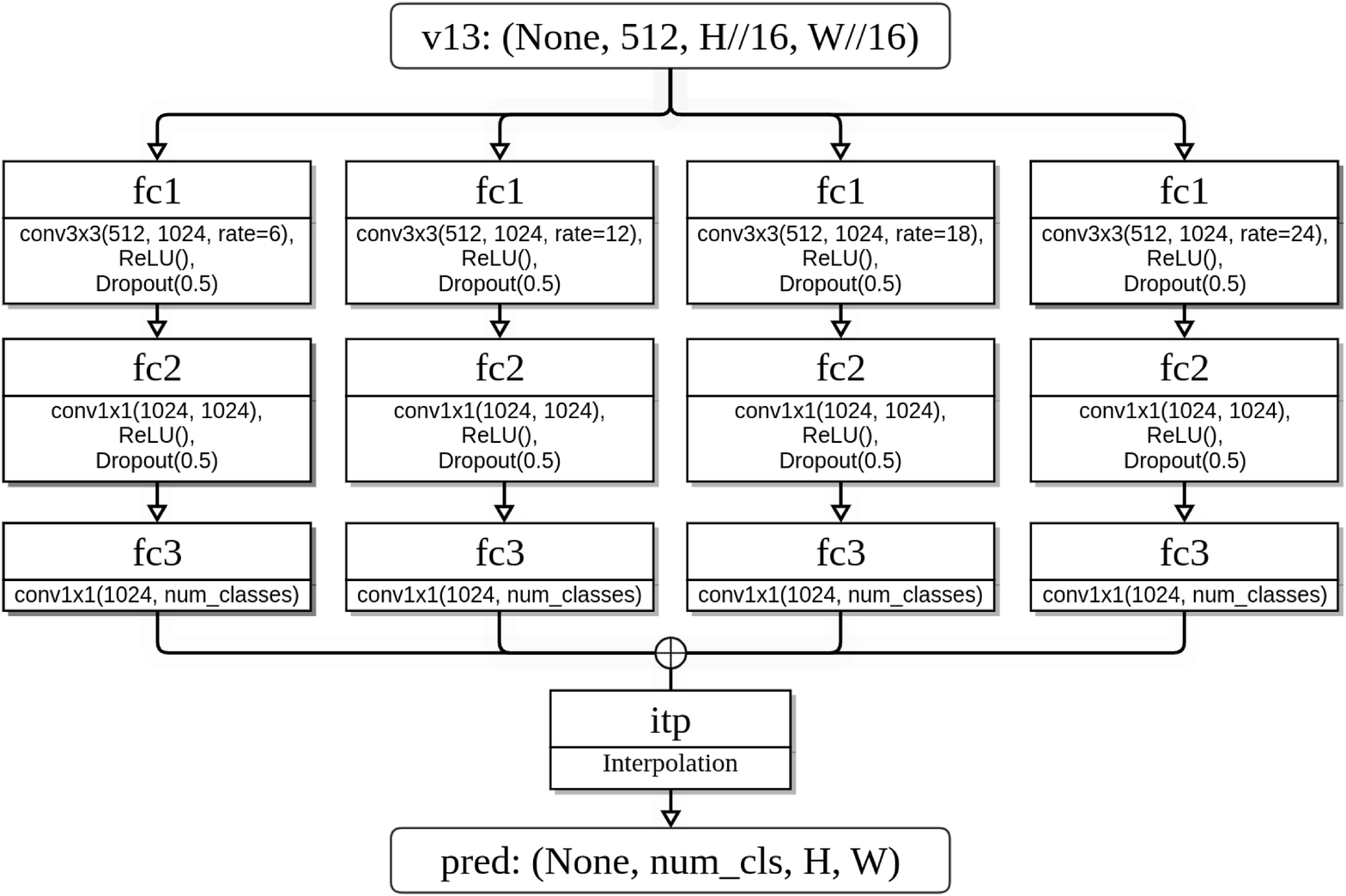}
  \caption{\textbf{Task-specific head configuration}}
  \label{fig:head-config}
\end{figure}

\noindent\textbf{Task-specific Head Architecture}
For NYU-v2~\cite{silberman2012indoor}, Cityscapes~\cite{Cordts2016Cityscapes}, and PASCAL-Context~\cite{mottaghi2014role}, we use the ASPP~\cite{chen2017deeplab} architecture, a popular architecture for pixel-wise prediction tasks, as our task-specific heads. 

\noindent\textbf{Training Details}
The overall training process of our framework consists of 3 stages: warm-up, search, and fine-tuning.
For Omniglot~\cite{lake2015human}, we train the network 2,000, 3,000, and 5,000 iterations for warm-up, search, and fine-tuning stages, respectively.
Similarly, for both NYU-v2~\cite{silberman2012indoor} and Cityscapes~\cite{Cordts2016Cityscapes}, we train the network 5,000, 15,000, and 20,000 iterations for warm-up, search, and fine-tuning stages, respectively.
For PASCAL-Context~\cite{mottaghi2014role}, the network is trained for 10,000, 20,000, and 30,000 iterations for the warm-up, search, and fine-tuning stages, respectively.
We train all baselines~\cite{misra2016cross,ruder2017sluice,gao2019nddr,liu2019end,ahn2019deep,sun2020adashare,guo2020learning,raychaudhuri2022controllable} with the same number of fine-tuning iterations for a fair comparison.
Before the fine-tuning stage, we rewind the model parameters to the parameters at the end of the warm-up stage.
We also report the learning rates of model weights parameters and upper-level parameters, and the balancing hyperparameter of squeeze loss $\mathcal{L}_{sq}$ in the \tabref{tab:hyperparameters}.

\section{Full Results of All Metrics}

In addition to the relative performance of all datasets (in the main paper), we report all the absolute task performance of NYU-v2, Cityscapes, and PASCAL-Context dataset with baseline in \tabref{tab:nyuv2}-\ref{tab:pascalcontext}.

\begin{figure}
  \centering
  \includegraphics[width=0.95\linewidth]{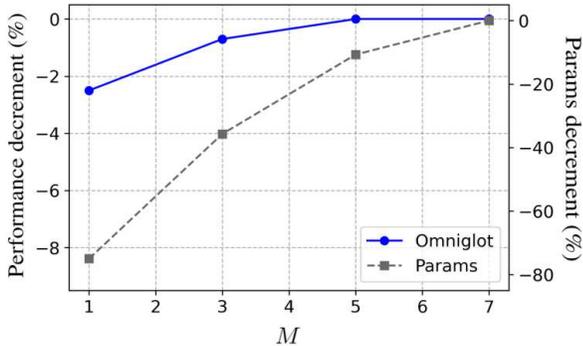}
  \caption{\textbf{Model performance with respect to the proposed flow-restriction (Omniglot)}}
  \label{fig:sup_chart1}
\end{figure}

\begin{figure}
  \centering
  \includegraphics[width=0.95\linewidth]{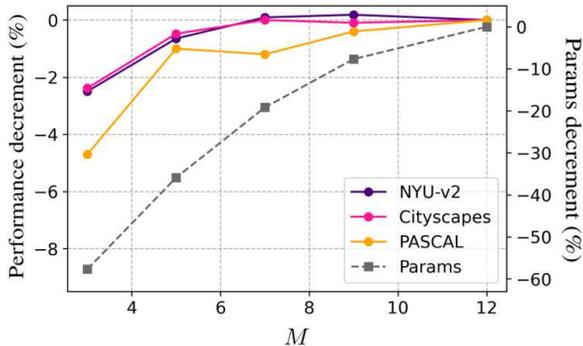}
  \caption{\textbf{Model performance with respect to the proposed flow-restriction (NYU-v2, Cityscapes, PASCAL-Context)}}
  \label{fig:sup_chart2}
\end{figure}

\section{Trade-off Curves of All Datasets}
Similar to Sec. 4.4 in the main paper, we analyze performance and computational complexity with respect to the flow constant $M$ for all datasets. 
We plot the degradation ratio of the performance (left y-axis) and parameter (right y-axis) by changing the flow constant $M$ in \figref{fig:sup_chart1}-\ref{fig:sup_chart2}.
The final task performance degradation of each dataset, including Omniglot, NYU-v2, Cityscapes, and PASCAL-Context, is marked by blue, purple, pink, and orange, respectively.
Additionally, the number of parameters of search space for Omniglot, and other datasets are marked by a gray dashed line. 

\begin{table}[t]
\centering
\resizebox{\linewidth}{!}{%
\begin{tabular}{c|ccc}

\toprule
\textbf{Dataset} & weight lr & upper lr & $\lambda_{sq}$\\
\midrule
Omniglot~\cite{lake2015human}          & 0.0001 & 0.01 & 0.05 \\
NYU-v2~\cite{silberman2012indoor}      & 0.0001 & 0.01 & 0.05 \\
Cityscapes~\cite{Cordts2016Cityscapes} & 0.0001 & 0.05 & 0.01 \\
PASCAL-Context~\cite{mottaghi2014role} & 0.0001 & 0.01 & 0.005 \\
\bottomrule
\end{tabular}
}
\caption{\textbf{Hyperparameters for each dataset} We report the learning rates of model weights parameters (weight lr), and upper-level parameters (upper lr). and balancing weight $\lambda_{sq}$ for squeeze loss $\mathcal{L}_{sq}$. \textbf{Our framework does not use task-balancing parameters.}}
\label{tab:hyperparameters}
\end{table}

\section{Ablation Studies}
\subsection{Three-stage learning scheme}
We follow the learning scheme as traditional Nas-style MTL three-stage learning.
To show that the three-stage learning scheme boosts the overall performance on multi-task learning scenarios, we report the relative task performance of each stage in \tabref{tab:threestage}.

\begin{table}[h]
\centering
\resizebox{\linewidth}{!}{%
\begin{tabular}{c|ccc|c|c}

\toprule
Method $(M=5)$ & $\Delta_{\mathcal{T}_{sem}} \uparrow$ & $\Delta_{\mathcal{T}_{dep}} \uparrow$ & $\Delta_{\mathcal{T}_{norm}} \uparrow$ & $\Delta_{\mathcal{T}} \uparrow$ & $\#$ of Param $\downarrow$\\
\midrule
with three-stages & \textbf{0.0} & \textbf{0.0} & \textbf{0.0} & \textbf{0.0} & \textbf{1.04} \\
\midrule
w/o warm-up & -7.4 & -3.7 & -3.0 & -4.3 & \textbf{1.04} \\
w/o search + FBR & -14.8 & -0.1 & -3.3 & -6.1 & 6.50 \\
w/o fine-tune & -13.6 & -9.7 & -3.3 & -8.9 & \textbf{1.04} \\
\bottomrule
\end{tabular}
}
\vspace{-3mm}
\caption{{\footnotesize\textbf{Ablation studies of three-stages on \textbf{NYU-v2} dataset}}}
\vspace{-5mm}
\label{tab:threestage}
\end{table}

\subsection{Ablation studies on key components}
Lastly, we provide the absolute task performance of all metrics for ablation studies of four key components; flow restriction, read-in/out layers, flow-based reduction, and squeeze loss in \tabref{tab:sup_ablation}.


\clearpage

\begin{table*}
\centering
\resizebox{0.9\linewidth}{!}{%
\begin{tabular}{c|c|cc|ccccc|ccccc}
\toprule
& & \multicolumn{2}{c}{\textbf{Semantic Seg.}}  & \multicolumn{5}{c}{\textbf{Depth Prediction}}  & \multicolumn{5}{c}{\textbf{Surface Normal Prediction}}  \\[0.5ex] 
\cline{3-14}
\textbf{Method}         & \textbf{$\#$ Params $\downarrow$} & \multirow{2}{*}{mIoU $\uparrow$} & \multirow{2}{*}{Pixel Acc $\uparrow$} & \multicolumn{2}{c}{Error $\downarrow$} & \multicolumn{3}{c|}{$\theta$, within $\uparrow$} & \multicolumn{2}{c}{Error $\downarrow$} & \multicolumn{3}{c}{$\delta$, within $\uparrow$}     \\[0.3ex]
\cline{5-14}
&              &                                &                                     & Abs               & Rel              & $1.25$       & $1.25^2$       & $1.25^3$      & Mean             & Median            & $11.25^\circ$ & $22.5^\circ$ & $30^\circ$ \\[0.3ex]
\midrule
Single-Task   & 3            & 27.5           & 58.9          & 0.62              & 0.25             & 57.9       & 85.8           & 95.7          & 17.5           & 15.2              & 34.9            & 73.3           & 85.7         \\
Shared Bottom & \textbf{1}            & 24.1           & 57.2          & 0.58              & 0.23             & 62.4       & 88.2           & 96.5          & 16.6           & 13.4              & 42.5            & 73.2           & 84.6         \\
Cross-Stitch  & 3            & 25.4           & 57.6          & 0.58              & 0.23             & 61.4       & 88.4           & 95.5          & 17.2           & 14.0              & 41.4            & 70.5           & 82.9         \\
Sluice        & 3            & 23.8           & 56.9          & 0.58              & 0.24             & 61.9       & 88.1           & 96.3          & 17.2           & 14.4              & 38.9            & 71.8           & 83.9         \\
NDDR-CNN      & 3.15         & 21.6           & 53.9          & 0.66              & 0.26             & 55.7       & 83.7           & 94.8          & 17.1           & 14.5              & 37.4            & 73.7           & 85.6         \\
MTAN          & 3.11         & 26.0           & 57.2          & 0.57              & 0.25             & 62.7       & 87.7           & 95.9          & 16.6           & 13.0              & 43.7            & 73.3           & 84.4         \\
DEN           & 1.12         & 23.9           & 54.9          & 0.97              & 0.31             & 22.8       & 62.4           & 88.2          & 17.1           & 14.8              & 36.0            & 73.4           & 85.9         \\
AdaShare      & \textbf{1}            & 30.2           & 62.4          & 0.55              & \textbf{0.20}             & 64.5       & 90.5           & 97.8          & 16.6             & \textbf{12.9}              & \textbf{45.0}            & 71.7           & 83.0       \\
\midrule
Ours $(M=5)$  & 1.04         & 31.8       & 63.7      & 0.56          & 0.21        & 64.3   & 90.2       & 97.7      
& 16.5             & 13.2              & 43.9            & 71.7           & 82.9       \\
Ours $(M=7)$  & 1.31         & \textbf{32.3}       & 64.3      & \textbf{0.54}          & \textbf{0.20}        & \textbf{64.7}   & 90.5       & 98.1      
& \textbf{16.4}             & \textbf{12.9}              & 43.1            & \textbf{73.8}           & \textbf{86.1}       \\
Ours $(M=9)$  & 1.63         & 32.1       & \textbf{64.6}      & \textbf{0.54}          & \textbf{0.20}        & \textbf{64.7}   & \textbf{91.1}       & \textbf{99.1}      
& \textbf{16.4}         & 13.1          &  43.4           & \textbf{73.8}       & 86.0   \\

\bottomrule
\end{tabular}
}
\vspace{-5pt}
\caption{\textbf{NYU v2 full results}}
\label{tab:nyuv2}
\vspace{-5pt}
\end{table*}

\begin{table*}
\label{tab:cityscape}
\centering
\resizebox{0.7\linewidth}{!}{%
\begin{tabular}{c|c|cc|ccccc}
\toprule
&  & \multicolumn{2}{c|}{\textbf{Semantic Seg.}} & \multicolumn{5}{c}{\textbf{Depth Prediction}}                       \\
[0.5ex]
\cline{3-9}
\textbf{Model}  &  \textbf{$\#$ Params $\downarrow$}  & \multirow{2}{*}{mIoU $\uparrow$} & \multirow{2}{*}{Pixel Acc $\uparrow$} & \multicolumn{2}{c}{Error $\downarrow$} & \multicolumn{3}{c}{$\delta$, within $\uparrow$} \\
\cline{5-9}
                       &         &             &            & Abs          & Rel        & 1.25   & $1.25^2$ & $1.25^3$ \\
\midrule
Single-Task            & 2       & 40.2        & 74.7       & 0.017        & 0.33       & 70.3   & 86.3     & 93.3     \\
Shared Bottom             & 1       & 37.7        & 73.8       & 0.018        & 0.34       & 72.4   & 88.3     & 94.2     \\
Cross-Stitch~\cite{misra2016cross}   & 2       & 40.3        & 74.3       & \textbf{0.015}        & \textbf{0.30}       & 74.2   & 89.3     & 94.9     \\
Sluice~\cite{ruder2017sluice}        & 2       & 39.8        & 74.2       & 0.016        & 0.31       & 73.0   & 88.8     & 94.6     \\
NDDR-CNN~\cite{gao2019nddr}               & 2.07    & 41.5        & 74.2       & 0.017        & 0.31       & 74.0   & 89.3     & 94.8     \\
MTAN~\cite{liu2019end}              & 2.41    & 40.8        & 74.3       & \textbf{0.015}        & 0.32       & 75.1   & 89.3     & 94.6     \\
DEN~\cite{ahn2019deep}         & 1.12    & 38.0        & 74.2       & 0.017        & 0.37       & 72.3   & 87.1     & 93.4     \\
AdaShare~\cite{sun2020adashare}               & 1       & 41.5        & 74.9       & 0.016        & 0.33       & \textbf{75.5}   & 89.8     & 94.9     \\
[0.5ex]
\midrule
Ours $(M=5)$            & \textbf{0.96}    & 42.8        & 75.1       & 0.016      & 0.32       & 74.8   & 89.1     & 94.2     \\
Ours $(M=7)$            & 1.16    & 46.4        & \textbf{75.6}       & 0.016      & 0.33       & 74.0   & 89.3     & 94.0     \\
Ours $(M=9)$            & 1.31    & \textbf{46.5}        & 75.4       & 0.016      & 0.32       & 75.4   & \textbf{90.4}     & \textbf{96.1}     \\
\bottomrule
\end{tabular}
}
\vspace{-5pt}
\caption{\textbf{Cityscapes full results}}
\vspace{-5pt}
\end{table*}

\begin{table*}
\centering
\resizebox{0.7\linewidth}{!}{%
\begin{tabular}{c|c|c|c|c|c|c}
\toprule
\multirow{2}{*}{\textbf{Method}} & \multirow{2}{*}{\textbf{$\#$ Params $\downarrow$}}   & \textbf{Semantic Seg.}    & \textbf{Part Seg.}  & \textbf{Saliency} &  \textbf{Surface Normal} & \textbf{Edge} \\
\cline{3-7}
 &  & mIoU $\uparrow$ & mIoU $\uparrow$ & mIoU $\uparrow$ & Mean $\downarrow$ & Mean $\downarrow$   \\
\midrule
Single-Task   & 5            & \textbf{63.9}           & 57.6          & 65.2              & 14.0             & \textbf{0.018}   \\
Shared Bottom & \textbf{1}            & 59.7           & 57.2          & 63.0              & 16.0             & \textbf{0.018}   \\
Cross-Stitch~\cite{misra2016cross}  & 5            & 63.1           & 59.7          & 65.1              & 14.2             & \textbf{0.018}   \\
Sluice~\cite{ruder2017sluice}        & 5            & 62.9           & 56.9          & 64.9              & 14.4             & 0.019   \\
NDDR-CNN~\cite{gao2019nddr}      & 5.61         & 63.2           & 56.1          & 65.2              & 14.7             & \textbf{0.018}   \\
MTAN~\cite{liu2019end}          & 5.21         & 61.6           & 57.2          & 65.0              & 14.7             & 0.019   \\
AdaShare~\cite{sun2020adashare}      & \textbf{1}            & 63.1           & \textbf{59.9}          & 64.9              & 14.1             & \textbf{0.018}   \\
LTB~\cite{guo2020learning} & 3.19            & 59.5           & 56.5          & 65.3              & 14.2             & \textbf{0.018}\\
PHN~\cite{raychaudhuri2022controllable} & 2.51            & 59.7           & 56.7          & 64.6         & 14.0       & \textbf{0.018}   \\
\midrule
Ours $(M=5)$ & 1.93  & 63.7 & 59.6  & 65.8 & 14.0 & \textbf{0.018}  \\
Ours $(M=7)$ & 1.91  & \textbf{63.9} & 57.5  & 66.3 & \textbf{13.8} & \textbf{0.018}  \\
Ours $(M=9)$ & 2.31  & \textbf{63.9} & 59.7  & \textbf{66.4} & \textbf{13.8} & \textbf{0.018}  \\
\bottomrule

\end{tabular}
}
\vspace{-5pt}
\caption{\textbf{PASCAL-Context full results}}
\label{tab:pascalcontext}
\vspace{-5pt}
\end{table*}

\begin{table*}
\centering
\resizebox{0.9\linewidth}{!}{%

\begin{tabular}{c|cc|ccccc|ccccc}
\toprule
& \multicolumn{2}{c}{\textbf{Semantic Seg.}}  & \multicolumn{5}{c}{\textbf{Depth Prediction}}  & \multicolumn{5}{c}{\textbf{Surface Normal Prediction}}  \\
\cline{2-13}
\textbf{Method} & \multirow{2}{*}{mIoU $\uparrow$} & \multirow{2}{*}{Pixel Acc $\uparrow$} & \multicolumn{2}{c}{Error $\downarrow$} & \multicolumn{3}{c|}{$\theta$, within $\uparrow$} & \multicolumn{2}{c}{Error $\downarrow$} & \multicolumn{3}{c}{$\delta$, within $\uparrow$}     \\
\cline{4-13}
&  &  & Abs    & Rel     & $1.25$   & $1.25^2$     & $1.25^3$   & Mean    & Median   & $11.25^\circ$ & $22.5^\circ$ & $30^\circ$ \\
\midrule
Ours $(M=7)$    & 32.3       & 64.3      & 0.54          & \textbf{0.20}        & 64.7   & 90.5       & 98.1      & \textbf{16.4}       & \textbf{12.9}      & \textbf{43.1}       & \textbf{73.8}         & 86.1       \\
\midrule
w/o flow-restriction     &32.1	&64.6	&0.54	&\textbf{0.20}	&64.2	&\textbf{90.7}	&98.1	&16.5	&\textbf{12.9}	&42.9	&73.7	&\textbf{87.2}\\
w/o read-in/out          &31.3	&64.5	&0.54	&\textbf{0.20}	&64.5	&90.3	&98.0	&16.6	&13.0	&42.5	&73.0	&86.3\\
w/o flow-based reduction &\textbf{32.5}	&\textbf{64.9}	&\textbf{0.53}	&\textbf{0.20}	&\textbf{64.8}	&\textbf{90.7}	&\textbf{98.3}	&\textbf{16.4}	&\textbf{12.9}	&\textbf{43.1}	&\textbf{73.8}	&86.3\\
w/o $\mathcal{L}_{sq}$   &32.1	&64.6	&0.54	&\textbf{0.20}	&64.7	&90.5	&98.1	&16.5	&13.0	&42.5	&73.6	&87.0\\
\bottomrule
\end{tabular}

}
\vspace{-5pt}
\caption{\textbf{Ablation Studies in NYU-v2}}
\label{tab:sup_ablation}
\end{table*}

\end{document}